\documentclass{article}

\usepackage{microtype}
\usepackage{graphicx}
\usepackage{subfigure}
\usepackage{booktabs} %
\usepackage{todonotes}
\usepackage{amssymb}
\usepackage{bm}
\usepackage{amsmath}
\usepackage{svg}
\usepackage{color}
\usepackage{soul}

\usepackage{hyperref}

\usepackage[accepted]{icml2018_style/icml2018}

\usepackage{xcolor}
\usepackage{colortbl}
\newcommand{\graycell}{\cellcolor{gray!25}}

\icmltitlerunning{MolGAN: An implicit generative model for small molecular graphs}

\begin{document}

\twocolumn[
\icmltitle{MolGAN: An implicit generative model for small molecular graphs}

\begin{icmlauthorlist}
\icmlauthor{Nicola De Cao}{ivi}
\icmlauthor{Thomas Kipf}{ivi}
\end{icmlauthorlist}

\icmlaffiliation{ivi}{Informatics Institute, University of Amsterdam, Amsterdam, The Netherlands}

\icmlcorrespondingauthor{Nicola De Cao}{\href{mailto:nicola.decao@gmail.com}{nicola.decao@gmail.com}}

\icmlkeywords{Machine Learning, ICML, DGM, GAN, RL}

\vskip 0.3in
]

\printAffiliationsAndNotice{}  %

\begin{abstract}
Deep generative models for graph-structured data offer a new angle on the problem of chemical synthesis: by optimizing differentiable models that directly generate molecular graphs, it is possible to side-step expensive search procedures in the discrete and vast space of chemical structures. We introduce MolGAN, an implicit, likelihood-free generative model for small molecular graphs that circumvents the need for expensive graph matching procedures or node ordering heuristics of previous likelihood-based methods. Our method adapts generative adversarial networks (GANs) to operate directly on graph-structured data. We combine our approach with a reinforcement learning objective to encourage the generation of molecules with specific desired chemical properties. In experiments on the QM9 chemical database, we demonstrate that our model is capable of generating close to $100\%$ valid compounds. \mbox{MolGAN} compares favorably both to recent proposals that use string-based (SMILES) representations of molecules and to a likelihood-based method that directly generates graphs, albeit being susceptible to mode collapse.
\end{abstract}

\section{Introduction}
\label{sec:introduction}

\begin{figure}[!tp]
    \centering
    \includegraphics[width=0.45\textwidth]{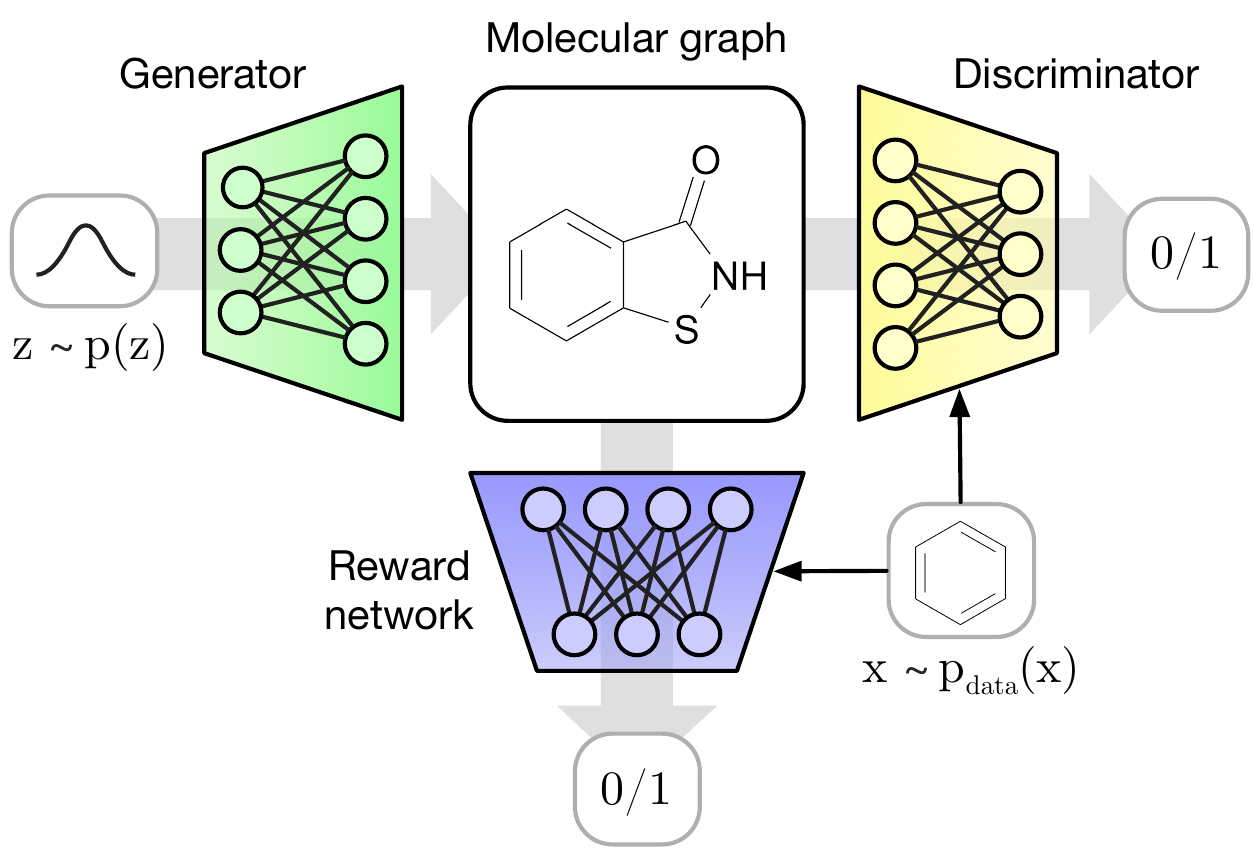}
    \caption{Schema of MolGAN. A vector $\bm{z}$ is sampled from a prior and passed to the generator which outputs the graph representation of a molecule. The discriminator classifies whether the molecular graph comes from the generator or the dataset. The reward network tries to estimate the reward for the chemical properties of a particular molecule provided by an external software.}
    \label{fig:MolGAN}
\end{figure}

Finding new chemical compounds with desired properties is a challenging task with important applications such as \emph{de novo} drug design \cite{schneider2005computer}. The space of synthesizable molecules is vast and search in this space proves to be very difficult, mostly owing to its discrete nature.

Recent progress in the development of deep generative models has spawned a range of promising proposals to address this issue. Most works in this area \cite{gomez2016automatic,kusner2017grammar,guimaraes2017objective,dai2018syntax} make use of a so-called SMILES representation \cite{weininger1988smiles} of molecules: a string-based representation derived from molecular graphs. Recurrent neural networks (RNNs) are ideal candidates for these representations and consequently, most recent works follow the recipe of applying RNN-based generative models on this type of encoding. String-based representations of molecules, however, have certain disadvantages: RNNs have to spend capacity on learning both the syntactic rules and the order ambiguity of the representation. 
Besides, this is approach not applicable to generic (non-molecular) graphs.
    
SMILES strings are generated from a graph-based representation of molecules, thereby working in the original graph space has the benefit of removing additional overhead. With recent progress in the area of deep learning on graphs \cite{bronstein2017geometric, hamilton2017representation}, training deep generative models directly on graph representations becomes a feasible alternative that has been explored in a range of recent works \cite{kipf2016variational,johnson2017learning,grover2018graphite,li2018learning,simonovsky2018graphvae,you2018graphrnn}. 
    
Likelihood-based methods for molecular graph generation \citep{li2018learning,simonovsky2018graphvae} however, either require providing a fixed (or randomly chosen) ordered representation of the graph or an expensive
graph %
matching procedure to evaluate the likelihood of a generated molecule, as the evaluation of all possible node orderings is prohibitive already for graphs of small size.

In this work, we sidestep this issue by utilizing implicit, likelihood-free methods, in particular, a generative adversarial network (GAN) \cite{goodfellow2014generative} that we adapt to work directly on graph representations. We further utilize a reinforcement learning (RL) objective similar to ORGAN \citep{guimaraes2017objective} to encourage the generation of molecules with particular properties. 

Our molecular GAN (MolGAN) model (outlined in Figure \ref{fig:MolGAN}) is the first to address the generation of graph-structured data in the context of molecular synthesis using GANs \cite{goodfellow2014generative}. The generative model of MolGAN predicts discrete graph structure at once (i.e., non-sequentially) for computational efficiency, although sequential variants are possible in general. MolGAN further utilizes a permutation-invariant discriminator and reward network (for RL-based optimization towards desired chemical properties) based on graph convolution layers \cite{bruna2013spectral,duvenaud2015convolutional,kipf2016semi,schlichtkrull2017modeling} that both operate directly on graph-structured representations.

\section{Background}

\subsection{Molecules as graphs} \label{sec:mol=graph}
Most previous deep generative models for molecular data \cite{gomez2016automatic,kusner2017grammar,guimaraes2017objective,dai2018syntax} resort to generating SMILES representations of molecules. The SMILES syntax, however, is not robust to small changes or mistakes, which can result in the generation of invalid or drastically different structures. Grammar VAEs \citep{kusner2017grammar} alleviate this problem by constraining the generative process to follow a particular grammar.

Operating directly in the space of graphs has recently been shown to be a viable alternative for generative modeling of molecular data \citep{li2018learning,simonovsky2018graphvae} with the added benefit that all generated outputs are valid graphs (but not necessarily valid molecules).

We consider that each molecule can be represented by an undirected graph $\mathcal{G}$ with a set of edges $\mathcal{E}$ and nodes $\mathcal{V}$. Each atom corresponds to a node $v_i\in\mathcal{V}$ that is associated with a $T$-dimensional one-hot vector $\mathbf{x}_i$, indicating the type of the atom. We further represent each atomic bond as an edge $(v_i,v_j)\in\mathcal{E}$ associated with a bond type $y\in\{1, ..., Y\}$. For a molecular graph with $N$ nodes, we can summarize this representation in a node feature matrix $\mathbf{X}=[\mathbf{x}_1, ..., \mathbf{x}_N]^T\in\mathbb{R}^{N\times T}$ and an adjacency tensor $\mathbf{A}\in\mathbb{R}^{N\times  N\times Y}$ where $\mathbf{A}_{ij}\in\mathbb{R}^{Y}$ is a one-hot vector indicating the type of the edge between $i$ and $j$.

\subsection{Implicit vs. likelihood-based methods}

Likelihood-based methods such as the variational auto-encoder (VAE) \cite{kingma2013auto,rezende2014stochastic} typically allow for easier and more stable optimization than implicit generative models such as a GAN \cite{goodfellow2014generative}. When generating graph-structured data, however, we wish to be invariant to reordering of nodes in the (ordered) matrix representation of the graph, which requires us to either perform a prohibitively expensive graph matching procedure \cite{simonovsky2018graphvae} or to evaluate the likelihood for all possible node permutations explicitly. 

By resorting to implicit generative models, in particular to the GAN framework, we circumvent the need for an explicit likelihood. While the discriminator of the GAN can be made invariant to node ordering by utilizing graph convolutions \cite{bruna2013spectral,duvenaud2015convolutional,kipf2016semi}
and a node aggregation operator \citep{li2016gated},
the generator still has to decide on a specific node ordering when generating a graph. Since we do not provide a likelihood, however, the generator is free to choose any suitable ordering for the task at hand. We provide a brief introduction to GANs in the following.

\paragraph{Generative adversarial networks} GANs \cite{goodfellow2014generative} are implicit generative models in the sense that they allow for inference of model parameters without requiring one to specify a likelihood.

A GAN consist of two main components: a generative model $G_\theta$, that learns a map from a prior to the data distribution to sample new data-points, and a discriminative model $D_\phi$, that learns to classify whether samples came from the data distribution rather than from $G_\theta$. Those two models are implemented as neural networks and trained simultaneously with stochastic gradient descent (SGD). $G_\theta$ and $D_\phi$ have different objectives, and they can be seen as two players in a minimax game
\begin{align}
    \min_\theta \max_\phi &\; \mathbb{E}_{\bm{x} \sim p_{data}(\bm{x}) }[\log D_\phi(\bm{x})] + \nonumber \\
    &\; \mathbb{E}_{\bm{z} \sim p_{\bm{z}}(\bm{z})}[\log (1 - D_\phi(G_\theta(\bm{z}))] \; ,
\end{align}
where $G_\theta$ tries to generate samples to fool the discriminator and $D_\phi$ tries to differentiate samples correctly. To prevent undesired behaviour such as mode collapse \citep{salimans2016improved} and to stabilize learning, we use minibatch discrimination \citep{salimans2016improved} and improved WGAN \citep{gulrajani2017improved}, an alternative and more stable GAN model that minimizes a better suited divergence.

\paragraph{Improved WGAN}
WGANs \citep{arjovsky2017wasserstein} minimize an approximation of the Earth Mover (EM) distance (also know as Wasserstein-1 distance) defined between two probability distributions. Formally, the Wasserstein distance between $p$ and $q$, using the Kantorovich-Rubinstein duality is
\begin{equation}
    D_W[p||q] = \frac1K \sup_{\|f\|_L < K} \mathbb{E}_{x\sim p(x)}\bigl[f(x)\bigl] - \mathbb{E}_{x\sim q(x)}\bigl[f(x)\bigl] \; ,
\end{equation}
where in the case of WGAN, $p$ is the empirical distribution and $q$ is the generator distribution. Note that the supremum is over all the K-Lipschitz functions for some $K>0$.

\citet{gulrajani2017improved} introduce a gradient penalty as an alternative soft constraint on the 1-Lipschitz continuity as an improvement upon the \emph{gradient clipping} scheme from the original WGAN. The loss with respect to the generator remains the same as in WGAN, but the loss function with respect to the discriminator is modified to be
\begin{align}
    L(\bm{x}^{(i)}, G_\theta(\bm{z}^{(i)});\phi) &= \underbrace{- D_\phi(\bm{x}^{(i)}) + D_\phi( G_\theta(\bm{z}^{(i)}) )}_{\text{original WGAN loss}} + \nonumber \\
    &\quad \underbrace{\alpha \left( \| \nabla_{\bm{\hat x}^{(i)}} D_\phi (\bm{\hat x}^{(i)}) \| - 1 \right)^2}_{\text{gradient penalty}} \;,
\end{align}
where $\alpha$ is a hyperparameter (we use $\alpha = 10$ as in the original paper), $\bm{\hat x}^{(i)}$ is a sampled linear combination between $\bm{x}^{(i)}\sim p_{data}(\bm{x})$ and $G_\theta(\bm{z}^{(i)})$ with $\bm{z}^{(i)}\sim p_{\bm{z}}(\bm{z})$, thus $\bm{\hat x}^{(i)} = \epsilon \; \bm{x}^{(i)} + (1 - \epsilon) \; G_\theta(\bm{z}^{(i)})$ with $\epsilon \sim \mathcal{U}(0,1)$.

\subsection{Deterministic policy gradients}

A GAN generator learns a transformation from a prior distribution to the data distribution. Thus, generated samples resemble data samples. However, in \emph{de novo} drug design methods, we are not only interested in generating chemically valid compounds, but we want them to have some useful property (e.g., to be easily synthesizable). Therefore, we also optimize the generation process towards some non-differentiable metrics using reinforcement learning.

In reinforcement learning,
a stochastic policy is represented by $\pi_\theta(s) = p_\theta(a|s)$ which is a parametric probability distribution in $\theta$ that selects a categorical action $a$ conditioned on an state $s$. Conversely, a deterministic policy is represented by $\mu_\theta(s) = a$ which deterministically outputs an action.

In initial experiments, we explored using REINFORCE \citep{williams1992simple} in combination with a stochastic policy that models graph generation as a set of categorical choices (actions). However, we found that it converged poorly due to the high dimensional action space when generating graphs at once. We instead base our method on a deterministic policy gradient algorithm which is known to perform well in high-dimensional action spaces \cite{silver2014deterministic}. In particular, we employ a simplified version of deep deterministic policy gradient (DDPG) introduced by \citet{lillicrap2015continuous}, an off-policy actor-critic algorithm that uses deterministic policy gradients to maximize an approximation of the expected future reward.

In our case, the policy is the GAN generator $G_\theta$ which takes a sample $\bm{z}$ for the prior as input, instead of an environmental state $s$, and it outputs a molecular graph as an action ($a=\mathcal{G}$). Moreover, we do not model episodes, so there is no need to assess the quality of a state-action combination 
since it does only depend on the graph $\mathcal{G}$. 
Therefore, we introduce a learnable and differentiable approximation of the reward function $\hat R_\psi(\mathcal{G})$ that predicts the immediate reward, and we train it via a mean squared error objective based on the real reward provided by an external system (e.g., the synthesizability score of a molecule). Then, we train the generator maximizing the predicted reward via $\hat R_\psi(\mathcal{G})$ which, being differentiable, provides a gradient
to the policy towards the desired metric.

\section{Model}

\begin{figure*}[!th]
    \centering
    \includegraphics[width=0.95\textwidth]{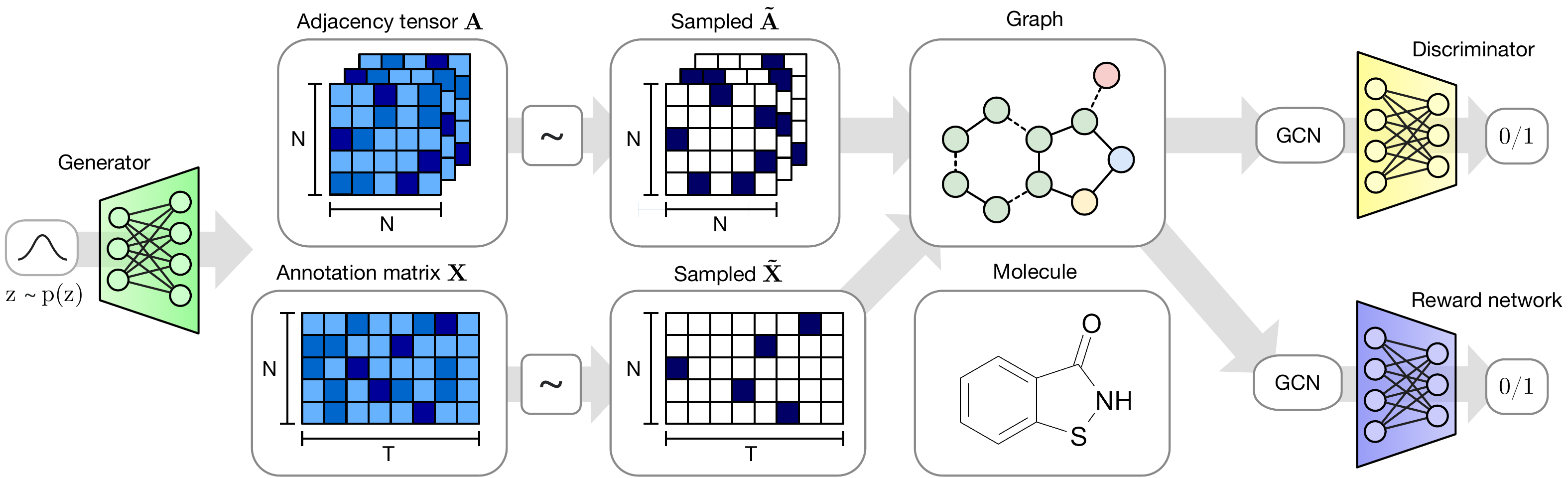}
    \caption{Outline of MolGAN. From \emph{left}: the generator takes a sample from a prior distribution and generates a dense adjacency tensor $\bm{A}$ and an annotation matrix $\bm{X}$. Subsequently, sparse and discrete $\bm{\tilde{A}}$ and $\bm{\tilde{X}}$ are obtained from $\bm{A}$ and $\bm{X}$ respectively via categorical sampling. The combination of $\bm{\tilde{A}}$ and $\bm{\tilde{X}}$ represents an annotated molecular graph which corresponds to a specific chemical compound. Finally, the graph is processed by both the discriminator and reward networks that are invariant to node order permutations and based on Relational-GCN \cite{schlichtkrull2017modeling} layers.}
    \label{fig:model}
\end{figure*}

The MolGAN architecture (Figure \ref{fig:model}) consists of three main components: a generator $G_\theta$, a discriminator $D_\phi$ and a reward network  $\hat R_\psi$.

The generator takes a sample from a prior distribution and generates an annotated graph $\mathcal{G}$ representing a molecule. Nodes and edges of $\mathcal{G}$ are associated with annotations denoting atom type and bond type respectively. The discriminator takes both samples from the dataset and the generator and learns to distinguish them. Both $G_\theta$ and $D_\phi$ are trained using improved WGAN such that the generator learns to match the empirical distribution and eventually outputs valid molecules. 

The reward network is used to approximate the reward function of a sample and optimize molecule generation towards non-differentiable metrics using reinforcement learning. Dataset and generated samples are inputs of $\hat R_\psi$, but, differently from the discriminator, it assigns scores to them (e.g., how likely the generated molecule is to be soluble in water). The reward network learns to assign a reward to each molecule to match a score provided by an external software\footnote{We used the RDKit Open-Source Cheminformatics Software: \url{http://www.rdkit.org}.}. Notice that, when MolGAN outputs a non-valid molecule, it is not possible to assign a reward since the graph is not even a compound. Thus, for invalid molecular graphs, we assign zero rewards.

The discriminator is trained using the WGAN objective while the generator uses a linear combination of the WGAN loss and the RL loss:
\begin{align} \label{equ:loss}
     L(\theta) = \lambda \cdot L_{WGAN}(\theta) + (1 - \lambda) \cdot L_{RL}(\theta) \; ,
\end{align}
where $\lambda \in [0,1]$ is a hyperparameter that regulates the trade-off between the two components. %

\subsection{Generator} \label{sec:generator}

$G_\phi(\bm{z})$ takes D-dimensional vectors $\bm{z} \in \mathbb{R}^D$ sampled from a standard normal distribution $\bm{z}\sim\mathcal{N}(\bm{0},\bm{I})$ and outputs graphs. While recent works have shown that it is feasible to generate graphs of small size by using an RNN-based generative model \citep{johnson2017learning, you2018graphrnn,li2018multi, li2018learning} we, for simplicity, utilize a generative model that predicts the entire graph at once using a simple multi-layer perceptron (MLP), as similarly done in \citet{simonovsky2018graphvae}. While this limits our study to graphs of a pre-chosen maximum size, we find that it is significantly faster and easier to optimize.

We restrict the domain to graphs of a limited number of nodes and, for each $\bm{z}$, $G_\theta$ outputs two continuous and dense objects: $\bm{X} \in \mathbb{R}^{N \times T}$ that defines atom types and $\bm{A}^{N \times N \times Y}$ that defines bonds types (see Section \ref{sec:mol=graph}).
Both $\bm{X}$ and $\bm{A}$ have a probabilistic interpretation since each node and edge type is represented with probabilities of categorical distributions over types. To generate a molecule we obtain discrete, sparse objects $\bm{\tilde{X}}$ and $\bm{\tilde{A}}$ via categorical sampling from $\bm{X}$ and $\bm{A}$, respectively. We overload notation and also represent samples from the dataset with binary  $\bm{\tilde{X}}$ and $\bm{\tilde{A}}$.

As this discretization process is non-differentiable, we explore three model variations to allow for gradient-based training:
We can i) use the continuous objects $\bm{X}$ and $\bm{A}$ directly during the forward pass (i.e., $\bm{\tilde{X}} = \bm{X}$ and $\bm{\tilde{A}} =\bm{A}$),  ii) add Gumbel noise to $\bm{X}$ and $\bm{A}$ before passing them to $D_\phi$ and $\hat R_\psi$ in order to make the generation stochastic while still forwarding continuous objects (i.e., $\bm{\tilde{X}}_{ij} = \bm{X}_{ij} + \mathrm{Gumbel}(\mu=0, \beta=1)$ and $\bm{\tilde{A}} =\bm{A}_{ijy} + \mathrm{Gumbel}(\mu=0, \beta=1)$), %
or iii) use a straight-through gradient based on categorical reparameterization with the Gumbel-Softmax \citep{jang2016categorical,maddison2016concrete}, that is we use a sample form a categorical distribution during the forward pass (i.e., $\bm{\tilde{X}}_i = \mathrm{Cat}(\bm{X}_i)$ and $\bm{\tilde{A}_{ij}} =\mathrm{Cat}(\bm{A}_{ij})$) and the continuous relaxed values (i.e., the original $\bm{X}$ and $\bm{A}$) in the backward pass.

\subsection{Discriminator and reward network}

Both the discriminator $D_\phi$ and the reward network $\hat R_\psi$ receive a graph as input, and they output a scalar value each. We choose the same architecture for both networks but do not share parameters between them. A series of graph convolution layers convolve node signals $\bm{\tilde{X}}$ using the graph adjacency tensor $\bm{\tilde{A}}$. We base our model on Relational-GCN \citep{schlichtkrull2017modeling}, a convolutional network for graphs with support for multiple edge types. At every layer, feature representations of nodes are convolved/propagated according to:
\begin{align} \label{equ:gcn}
    \bm{h}'^{(\ell + 1)}_i &= f_s^{(\ell)}(\bm{h}^{(\ell)}_i,\bm{x}_i) + \sum_{j=1}^N \sum_{y = 1}^Y \frac{\bm{\tilde{A}}_{ijy}}{|\mathcal{N}_i|} f_y^{(\ell)}(\bm{h}_j^{(\ell)},\bm{x}_j)\;, \nonumber \\
     \bm{h}^{(\ell + 1)}_i &=  \tanh ( \bm{h}'^{(\ell + 1)}_i ) \;,
\end{align}
where  $\bm{h}^{(\ell)}_i$ is the signal of the node $i$ at layer $\ell$ and $f_s^{(\ell)}$ is a linear transformation function that acts as a self-connection between layers. We further utilize an edge type-specific affine function $f_y^{(\ell)}$ for each layer. $\mathcal{N}_i$ denotes the set of neighbors for node $i$. The normalization factor $1/|\mathcal{N}_i|$ ensures that activations are on a similar scale irrespective of the number of neighbors. %

After several layers of propagation via graph convolutions, following \citet{li2016gated} we aggregate node embeddings into a graph level representation vector as
\begin{align} \label{equ:aggregation}
    \bm{h}'_{\mathcal{G}} &= \sum_{v \in \mathcal{V}} \sigma (i  (\bm{h}^{(L)}_v, \bm{x}_v ) ) \odot \tanh (j  (\bm{h}^{(L)}_v, \bm{x}_v ) ) \;,  \nonumber \\
    \bm{h}_{\mathcal{G}} &= \tanh\bm{h}'_{\mathcal{G}} \;,
\end{align}
where $\sigma(x)=1/(1+\exp(-x))$ is the logistic sigmoid function, $i$ and $j$ are MLPs with a linear output layer and $\odot$ denotes element-wise multiplication. Then, $\bm{h}_{\mathcal{G}}$ is a vector representation of the graph $\mathcal{G}$ and it is further processed by an MLP to produce a graph level scalar output $\in (-\infty, +\infty)$ for the discriminator and $\in (0,1)$ for the reward network.

\section{Related work}

Objective-Reinforced Generative Adversarial Networks (ORGAN) by \citet{guimaraes2017objective} is the closest related work to ours. Their model relies on SeqGAN \citep{yu2017seqgan} to adversarially learn to output sequences while optimizing towards chemical metrics with REINFORCE \citep{williams1992simple}. The main differences from our approach is that they model sequences of SMILES as molecular representations instead of graphs, and their RL component uses REINFORCE while we use DDPG. \citet{segler2017learning}
also employs RL for drug discovery by searching retrosynthetic routes using Monte Carlo Tree Search (MCTS) in combination with an expansion policy network.

Several other works have explored training generative models on SMILES representations of molecules: CharacterVAE \cite{gomez2016automatic} is the first such model that is based on a VAE with recurrent encoder and decoder networks. GrammarVAE \cite{kusner2017grammar} and SDVAE \cite{dai2018syntax} constrain the decoding process to follow particular syntactic and semantic rules.

A related line of research considers training deep generative models to output graph-structured data directly. Several works explored auto-encoder architectures utilizing graph convolutions for link prediction within graphs \cite{kipf2016variational,grover2018graphite,davidson2018hyperspherical}. \citet{johnson2017learning,li2018learning,you2018graphrnn,li2018multi} on the other hand developed likelihood-based methods to directly output graphs of arbitrary size in a sequential manner. Several related works have explored extending VAEs to generate graphs directly, examples include the GraphVAE \cite{simonovsky2018graphvae}, Junction Tree VAE \citep{jin2018junction} and the NeVAE \citep{samanta2018designing} model. %

For link prediction within graphs, a range of adversarial methods have been introduced in the literature \cite{minervini2017adversarial,wang2017graphgan,bojchevski2018netgan}. This class of models, however, is not suitable to generate molecular graphs from scratch, which makes direct comparison infeasible.

\section{Experiments}

We compare MolGAN against recent neural network-based drug generation models in a range of experiments on established benchmarks using the QM9 \citep{ramakrishnan2014quantum} chemical database. We first focus on studying the effect of the $\lambda$ parameter to find the best trade-off between the GAN and RL objective (see Section \ref{sec:lambda}). We then compare MolGAN with ORGAN \citep{guimaraes2017objective} since it is the most related work to ours: ORGAN is a sequential generative model operating on SMILES representations, optimizing towards several chemical properties with an RL objective (see Section \ref{sec:objectives}). We also compare our model against variational autoencoding methods  (Section \ref{sec:vae}) such as CharacterVAE \cite{gomez2016automatic}, GrammarVAE \cite{kusner2017grammar}, as well as a recent graph-based generative model: GraphVAE \cite{simonovsky2018graphvae}.

\paragraph{Dataset}
In all experiments, we used QM9 \citep{ramakrishnan2014quantum} a subset of the massive 166.4 billion molecules GDB-17 chemical database \citep{ruddigkeit2012enumeration}. QM9 contains 133,885 organic compounds up to 9 heavy atoms: carbon (C), oxygen (O), nitrogen (N) and fluorine (F). 

\paragraph{Generator architecture}
The generator architecture is fixed for all experiments. We use $N=9$ as the maximum number of nodes, $T=5$ as the number of atom types (C, O, N, F, and one padding symbol), and $Y=4$ as the number of bond types (single, double, triple and no bond). These dimensionalities are enough to cover all molecules in QM9. The generator takes a 32-dimensional vector sampled from a standard normal distribution $\bm{z}\sim\mathcal{N}(\bm{0},\bm{I})$ and process it with a 3-layer MLP of $[128,256,512]$ hidden units respectively, with $\tanh$ as activation functions. Eventually, the last layer is linearly projected to match $\bm{X}$ and $\bm{A}$ dimensions and normalized in their last dimension with a $\mathrm{softmax}$ operation ($\mathrm{softmax}(\bm{x})_i =  \exp(\bm{x_i}) / \sum_{i=1}^D \exp(\bm{x}_i)$).

\paragraph{Discriminator and reward network architecture}
Both networks use a RelationalGCN encoder (see Eq.~\ref{equ:gcn}) with two layers and $[64,32]$ hidden units, respectively, to process the input graphs. Subsequently, we compute a 128-dimensional graph-level representation (see Eq.~\ref{equ:aggregation}) further processed by a 2-layer MLP of dimensions $[128, 1]$ and with $\tanh$ as hidden layer activation function. In the reward network, we further use a sigmoid activation function on the output.

\paragraph{Evaluation measures}
We measure the following statistics as defined in \citet{samanta2018designing}: validity, novelty, and uniqueness. \emph{Validity} is defined as the ratio between the number of valid and all generated molecules. \emph{Novelty} measures the ratio between the set of valid samples that are not in the dataset and the total number of valid samples. Finally, \emph{uniqueness} is defined as the ratio between the number of unique samples and valid samples and it measures the degree of variety during sampling.

\paragraph{Training}
In all experiments, we use a batch size of 32 and train using the Adam \citep{kingma2014adam} optimizer with a learning rate of $10^{-3}$. For each setting, we employ a grid search over dropout rates $\in \{0.0, 0.1, 0.25\}$ \citep{srivastava2014dropout} as well as over discretization variations (as described in Section \ref{sec:generator}). We always report the results of the best model depending on what we are optimizing for (e.g., when optimizing solubility we report the model with the highest solubility score -- when no metric is optimized we report the model with the highest sum of individual scores).
Although the use of WGAN should prevent, to some extent, undesired behaviors like \emph{mode collapse} \citep{salimans2016improved}, we notice that our models suffer from that problem. We leave addressing this issue for future work. As a simple countermeasure, we employ early stopping, evaluating every 10 epochs, to avoid completely collapsed modes. In particular, we use the unique score to measure the degree of collapse of our models since it intrinsically indicates how much variety there is in the generation process. We set an arbitrary threshold of 2\% under which we consider a model to be collapsed and stop training.

During early stages of our work, we noticed that the reward network needs several epochs of pretraining before being used to propagate the gradient to the generator, otherwise the generator easily diverges. We think this happens because at the beginning of the training, $\hat R_\psi$ does not predict the reward accurately and then it does not optimize the generator well. Therefore, in each experiment, we train the generator for the first half of the epochs without the RL component, but using the WGAN objective only. We train the reward network during these epochs, but no RL loss is used to train the generator. For the second half of the epochs we use the combined loss in Equation \ref{equ:loss}.

\subsection{Effect of $\lambda$} \label{sec:lambda}

As in \citet{guimaraes2017objective}, the $\lambda$ hyperparameter controls the trade-off between maximizing the desired objective and regulating the generator to match the data distribution. We study the effects of $\lambda \in \{0.0, 0.01, 0.05, 0.5, 1.0\}$ on the solubility metric (see Section \ref{sec:objectives} for more details). %
We train for 300 epochs (150 of which for pretraining) on the 5k subset of QM9 used in \citet{guimaraes2017objective}. We use the best $\lambda$ parameter -- determined via the model with the maximum sum of valid, unique, novel, and solubility scores -- on all other experiments (Section \ref{sec:objectives} and \ref{sec:vae}) without doing any further search. %

\paragraph{Results}
We report results in Table \ref{tab:lambda}. We observe a clear trend towards higher validity scores for lower values of $\lambda$. This is likely due to the implicit optimization of valid molecules since invalid ones receive zero reward during training. Therefore, if the RL loss component is strong, the generator is optimized to generate mostly valid molecular graphs.
Conversely, it appears that $\lambda$ does not mainly affect the unique and novel scores. Notice that these scores are not optimized, neither directly nor indirectly, and therefore they are a result of model architecture, hyperparameters, and training procedure. Indeed, the unique score is always close to 2\% (which is our threshold) indicating that models appear to collapse (even in the RL only case) if we do not apply early stopping.

We also run $\lambda=0$ without starting from a pretrained model. We observe that  it succeeds in optimizing toward the desired metrics, but it collapses outputting very few samples (i.e., low unique score). This behavior may indicate that pretraining is fundamental for matching the data distribution before using RL since the GAN acts regularizing towards diversity.

Since $\lambda$ controls the trade-off between the WGAN and RL losses, it is not surprising that $\lambda=0$ (i.e., only RL in the second half of training) results in the highest valid and solubility scores compared to other values. The $\lambda$ value with the highest sum of scores is $\lambda=0$. We use this value for subsequent experiments.

\begin{table}[!ht]
\centering
\begin{tabular}{lcccc}
\toprule
Algorithm & Valid & Unique & Novel & Sol. \\
\midrule
$\lambda=0$ (full RL)* & 100.0 & \underline{0.03} & 100.0 & 0.98 \\
$\lambda=0$ (full RL) & \textbf{99.8} & 2.3 & 97.9 & \textbf{0.86} \\
$\lambda=0.01$ & 98.2 & 2.2 & \textbf{98.1} & 0.74 \\
$\lambda=0.05$ & 92.2 & 2.7 & 95.0 & 0.67 \\
$\lambda=0.1$ & 87.3 & \textbf{3.2} & 87.2 & 0.56 \\
$\lambda=0.5$ & 86.6 & 2.1 & 87.5 & 0.48 \\
$\lambda=1$ (no RL) & 87.7 & 2.9 & 97.7 & 0.54 \\
\bottomrule
\end{tabular}
\caption{Comparison of different combinations of RL and GAN objectives on the small 5k dataset after GAN-based  pretraining for 150 epochs. All values are reported in percentages except for the solubility score. * indicates no GAN-based pretraining and \emph{Sol.} indicates Solubility.}
\label{tab:lambda}
\end{table}

\subsection{Objectives optimization} \label{sec:objectives}

\begin{table*}[!ht]
\centering
\resizebox{\linewidth}{!}{%
\begin{tabular}{llccccccc}
\toprule
Objective & Algorithm & Valid (\%) &  Unique (\%) & Time (h) & Diversity & Druglikeliness & Synthesizability & Solubility  \\
\midrule
Druglikeliness & ORGAN & 88.2 & 69.4* & 9.63* & 0.55 & 0.52 \graycell & 0.32 &0.35 \\
& OR(W)GAN & 85.0 & 8.2* & 10.06* & 0.95 & \graycell 0.60 & 0.54 &0.47 \\
& Naive RL & 97.1 & 54.0* & 9.39* & 0.80 & \graycell  0.57 & 0.53 &0.50 \\
& \emph{MolGAN} & \textbf{99.9} & 2.0 & 1.66 & 0.95 & \graycell \textbf{0.61} & 0.68 & 0.52 \\
& \emph{MolGAN (QM9)} & \textbf{100.0} & 2.2 & 4.12 & \textbf{0.97} & \graycell \textbf{0.62} & 0.59 & 0.53 \\
\midrule
Synthesizability & ORGAN & 96.5 & 45.9* & 8.66* & 0.92 & 0.51 & \graycell 0.83  & 0.45 \\
& OR(W)GAN & 97.6 & 30.7* & 9.60* & \textbf{1.00} & 0.20 & \graycell 0.75  & 0.84 \\
& Naive RL & 97.7 & 13.6* & 10.60* & 0.96 & 0.52 & \graycell  0.83  & 0.46 \\
& \emph{MolGAN} & \textbf{99.4} & 2.1 & 1.04 & 0.75 & 0.52 & \graycell \textbf{0.90} & 0.67 \\
& \emph{MolGAN (QM9)} & \textbf{100.0} & 2.1 & 2.49 & 0.95 & 0.53 & \graycell \textbf{0.95} & 0.68 \\
\midrule
Solubility & ORGAN & 94.7 & 54.3* & 8.65* & 0.76 & 0.50 & 0.63 & \graycell 0.55 \\
& OR(W)GAN & 94.1 & 20.8* & 9.21* & 0.90 & 0.42 & 0.66 & \graycell 0.54 \\
& Naive RL & 92.7 & 100.0* & 10.51* & 0.75 & 0.49 & 0.70 & \graycell 0.78 \\
& \emph{MolGAN} & \textbf{99.8} & 2.3 & 0.58 & \textbf{0.97} & 0.45 & 0.42 & \graycell \textbf{0.86} \\
& \emph{MolGAN (QM9)} & \textbf{99.8} & 2.0 & 1.62 & \textbf{0.99} & 0.44 & 0.22 & \graycell \textbf{0.89} \\
\midrule
All/Alternated & ORGAN & 96.1 & 97.2* & 10.2* & 0.92 & \graycell \textbf{0.52} & \graycell 0.71 & \graycell 0.53 \\
All/Simultaneously & \emph{MolGAN} & \textbf{97.4} & 2.4 & 2.12 & 0.91 & \graycell 0.47 & \graycell \textbf{0.84} & \graycell \textbf{0.65} \\
All/Simultaneously & \emph{MolGAN (QM9)} & \textbf{98.0} & 2.3 & 5.83 & \textbf{0.93} & \graycell 0.51 & \graycell \textbf{0.82} & \graycell \textbf{0.69} \\
\bottomrule
\end{tabular}
}
\caption{Gray cells indicate directly optimized objectives. Baseline results are taken from \citet{guimaraes2017objective} (Table 1) and * indicates results reproduced by us using the code provided by the authors.}
\label{tab:results}
\end{table*}

Similarly to the previous experiment, we train our model for 300 epochs on the 5k QM9 subset while optimizing the same objectives as \citet{guimaraes2017objective} to compare against their work. Moreover, we also report results on the full dataset trained for 30 epochs (note that the full dataset is 20 times larger than the subset). All scores are normalized to lie within $[0,1]$. We assign a score of zero to invalid compounds (i.e., implicitly we are also optimizing a validity score). We choose to optimize the following objectives which represent qualities typically desired for drug discovery:

\paragraph{Druglikeness:} how likely a compound is to be a drug. The Quantitative Estimate of Druglikeness (QED) score quantifies compound quality with a weighted geometric mean of desirability scores capturing the underlying data distribution of several drug properties \cite{bickerton2012quantifying}.\vspace{-1em}
\paragraph{Solubility:} the degree to which a molecule is hydrophilic. The log octanol-water partition coefficient (logP), is defined as the logarithm of the ratio of the concentrations between two solvents of a solute \citep{comer2001lipophilicity}. \vspace{-1em} %
\paragraph{Synthetizability:} this measure quantifies how easy a molecule is to synthesize. The Synthetic Accessibility score \cite{ertl2009estimation} is a method to estimate the ease of synthesis in a probabilistic way.

We also measure, without optimizing for it, a diversity score which indicates how likely a molecule is to be diverse with respect to the dataset. This measure compares sub-structures between samples and a random subset from the dataset indicating how many repetitions there are. 

For evaluation, we report average scores from 6400 sampled compounds as in \citep{guimaraes2017objective}. Additionally, we re-run experiments from \citep{guimaraes2017objective} to compute unique scores and execution time since it is not reported. Differently from ORGAN, to optimize for all objectives, we do not alternate between optimizing them individually during training which in our case is not possible since the reward network is specific to a single type of reward. Thus, we instead optimize a joint reward which we define as the product (to lie within $\in [0,1]$) of all objectives.

\paragraph{Results}
Results are reported in Table \ref{tab:results}. Qualitative samples are provided in the Appendix (Figure \ref{fig:bestqed}). We observe that MolGAN models always converge to very high validity outputs $>97\%$ at the end of the training. This is coherent as observed in the previous experiment, since also here there is an implicit optimization of validity. Moreover, in all single metrics settings, our models beat ORGAN models in terms of valid scores as well as all the three objective scores we optimize for. 

We argue that this should be mainly due to two factors: i) intuitively, it should be easier to optimize a molecular graph predicted as a single sample than to optimize an RNN model that outputs a sequence of characters, and ii) using the deterministic policy gradient instead of REINFORCE
effectively provides a better gradient and it improves the sampling procedure towards metrics while penalizing invalid graphs.

Training on the full QM9 dataset for 10 times fewer epochs further improves results in almost all scores. During training, our algorithm observes more different samples, and therefore it can learn well with much fewer iterations. Moreover, it can observe molecules with more diverse structures and properties. %

As previously observed in Section \ref{sec:lambda}, also in this experiment the unique score is always close to 2\% confirming our hypothesis that our models are susceptible to \emph{mode collapse}. This is not the case for the ORGAN baseline. During sampling, ORGAN generates sequences of maximum 51 characters which allows it to generate larger molecules whereas our model is (by choice) constrained to generate up to 9 atoms. This explain the difference in unique score since the chance of generating different molecules in a smaller space is much lower. Notice that in ORGAN, the RL component relies on REINFORCE, and the unique score is optimized penalizing non-unique outputs which we do not.

In terms of training time, our model outperforms ORGAN by a large margin when training on the 5k dataset (at least $\sim$5 times faster in each setting), as we do not rely on sequential generation or discrimination. Both ORGAN and MolGAN have a comparable number of parameters, with the latter being approximately $20\%$ larger. %

\subsection{VAE Baselines} \label{sec:vae}

\begin{table}[tp]
\centering
\begin{tabular}{lccc}
\toprule
Algorithm & Valid & Unique & Novel \\
\midrule
CharacterVAE &  10.3 & 67.5 & 90.0 \\
GrammarVAE  &  60.2  & 9.3 & 80.9 \\
GraphVAE &  55.7 & \textbf{76.0} & 61.6 \\
GraphVAE/imp & 56.2 & 42.0 & 75.8 \\
GraphVAE NoGM &  81.0 & 24.1 & 61.0 \\
\midrule
MolGAN & \textbf{98.1} & 10.4 & \textbf{94.2} \\
\bottomrule
\end{tabular}
\caption{Comparison with different algorithms on QM9. Values are reported in percentages. Baseline results are taken from \citet{simonovsky2018graphvae}.}
\label{tab:vae}
\end{table}

In this experiment, we compare MolGAN against recent likelihood-based methods that utilize VAEs. We report a comparison with CharacterVAE \cite{gomez2016automatic}, GrammarVAE \cite{kusner2017grammar}, and GraphVAE \cite{simonovsky2018graphvae}. Here we train using the complete QM9 dataset. Naturally, we compare only with metrics that measure the quality of the generative process since the likelihood is not computed directly in MolGAN. Moreover, we do not optimize any particular chemical property except validity (i.e., we do not optimize any metric described above, but we optimize towards chemically valid compounds). The final evaluation scores are an average from $10^4$ random samples. The number of samples differs from the previous experiment to be in line with the setting in \citet{simonovsky2018graphvae}.

\paragraph{Results}
Results are reported in Table \ref{tab:vae}. Training on the full QM9 dataset (without optimizing any metric except validity) results in a model with a higher unique score compared to the ones in Section \ref{sec:objectives}. 

Though the unique score of MolGAN is slightly higher compared to GrammarVAE, the other baselines are superior in terms of this score. Even though here we do not consider our model to be collapsed, such a low score confirms our hypothesis that our model is prone to mode collapse. On the other hand, we observe significantly higher validity scores compared to the VAE-based baselines.%
To verify that sampled unique molecules are (mostly) novel and not simply memorized from the dataset, we additionally measure how many of the \emph{unique} molecules are also novel for our model. This score is 97\% indicating that almost all unique molecules are indeed novel and MolGAN does not suffer from such problems.

Differently from our approach, VAEs optimize the evidence lower bound (ELBO) and there is no explicit nor implicit optimization of output validity. Moreover, since a part of the ELBO maximizes reconstruction of the observations, the novelty in the sampling process is not expected to be high since it is not optimized. However, in all reported methods novelty is $>60\%$ and, in the case of CharacterVAE, $90\%$. Though CharacterVAE can achieve a high novelty score, it underperforms in terms of validity. MolGAN, on the other hand, achieves both high validity and novelty scores.

\section{Conclusions}
In this work, we have introduced MolGAN: an implicit generative model for molecular graphs of small size. Through joint training with a GAN and an RL objective, our model is capable of generating molecular graphs with both higher validity and novelty than previous comparable VAE-based generative models, while not requiring a permutation-dependent likelihood function. Compared to a recent SMILES-based sequential GAN model for molecular generation, MolGAN can achieve higher chemical property scores (such as solubility) while allowing for at least $\sim$5x faster training time.

A central limitation of our current formulation of MolGANs is their susceptibility to mode collapse: both the GAN and the RL objective do not encourage generation of diverse and non-unique outputs whereby the model tends to be pulled towards a solution that only involves little sample variability. This ultimately results in the generation of only a handful of different molecules if training is not stopped early.

We think that this issue can be addressed in future work, for example via careful design of reward functions or some form of pretraining. The MolGAN framework taken together with established benchmark datasets for chemical synthesis offer a new test bed for improvements on GAN stability with respect to the issue of mode collapse. We believe that insights gained from such evaluations will be valuable to the community even outside of the scope of generating molecular graphs. Lastly, it will be promising to explore alternative generative architectures within the MolGAN framework, such as recurrent graph-based generative models \cite{johnson2017learning,li2018learning,you2018graphrnn}, as our current one-shot prediction of the adjacency tensor is most likely feasible only for graphs of small size.

\section*{Acknowledgements}
The authors would like to thank Luca Falorsi, Tim R. Davidson, Herke van Hoof and Max Welling for helpful discussions and feedback. T.K.~is supported by SAP SE Berlin.

\bibliography{main}
\bibliographystyle{icml2018_style/icml2018}

\newpage
\appendix

\begin{figure*}[htp]
\centering

\subfigure[QM9 samples]{\includegraphics[width=0.45\textwidth]{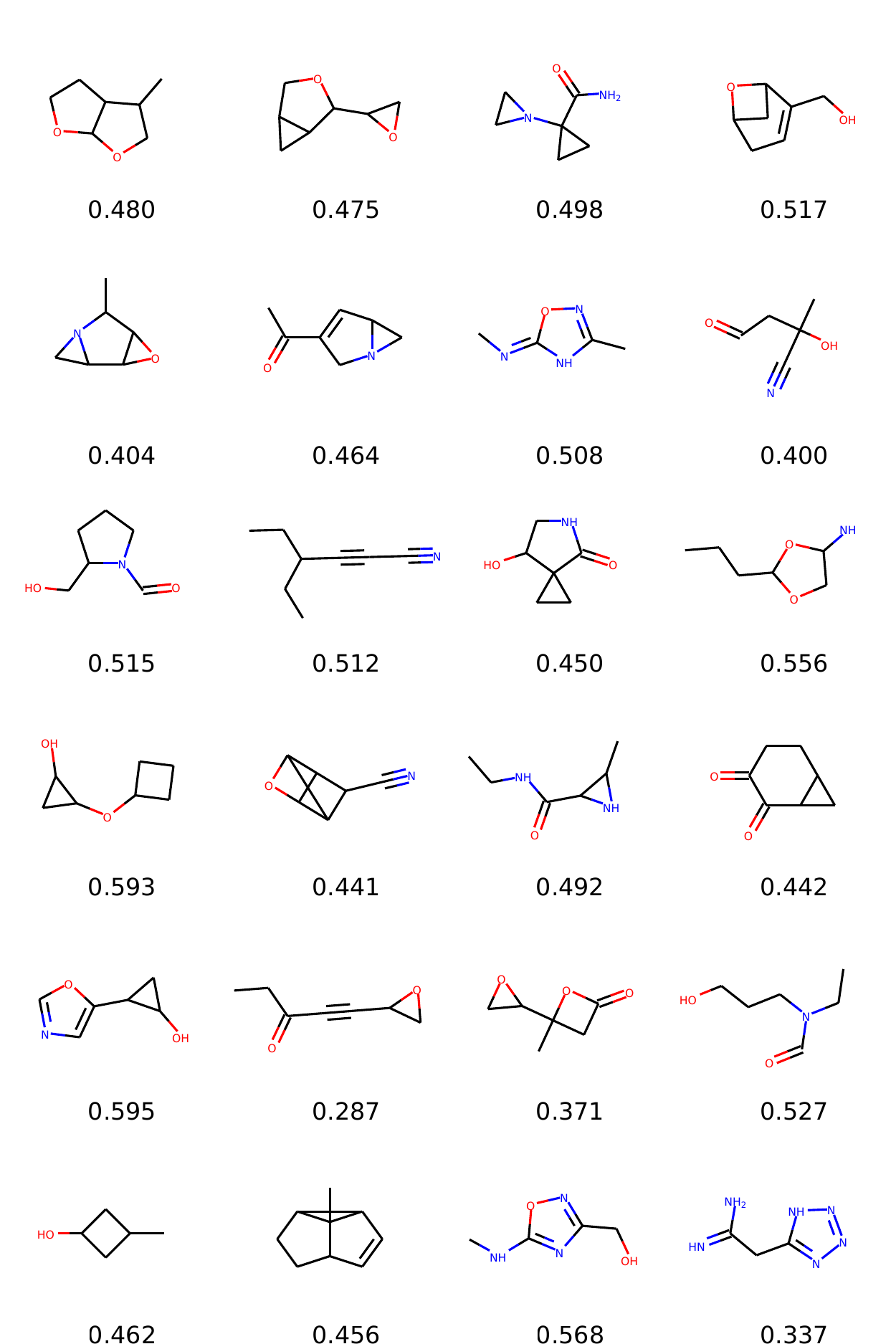}}
\hspace{1cm}
\subfigure[MolGAN (QED) samples]{\includegraphics[width=0.45\textwidth]{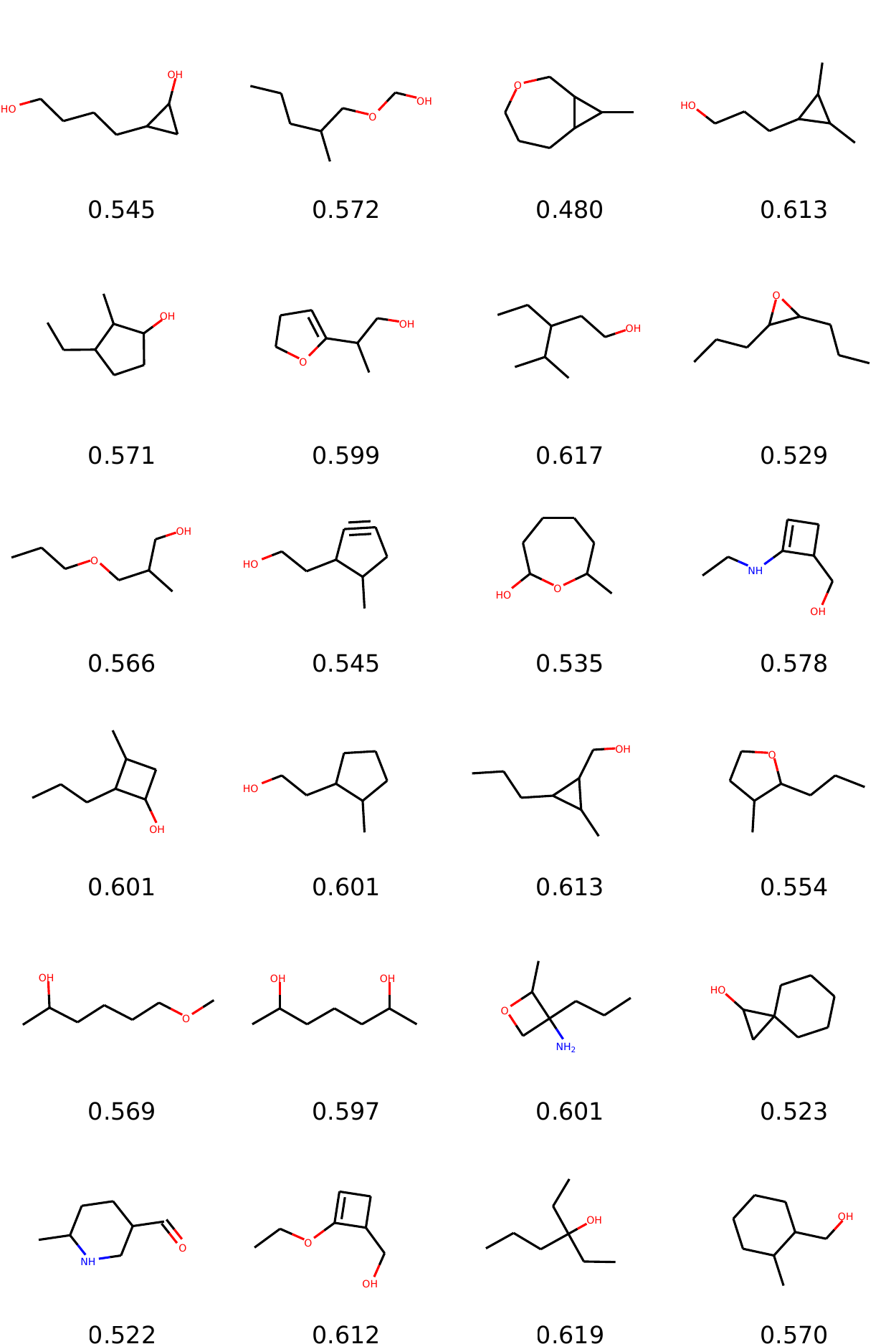}}
\caption{Samples from the QM9 dataset (\textit{left}) and MolGAN trained to optimize druglikeliness (QED) on the 5k QM9 subset (\textit{right}). We also report their relative QED scores.}
\label{fig:bestqed}
\end{figure*}

\end{document}